\documentclass{article}

\usepackage{arxiv}

\usepackage{microtype}
\usepackage{graphicx}
\usepackage{subcaption}
\usepackage{booktabs} 

\usepackage{amsmath,amsfonts,bm}

\def\eqref#1{equation~\ref{#1}}

\def\1{\bm{1}}

\DeclareMathAlphabet{\mathsfit}{\encodingdefault}{\sfdefault}{m}{sl}
\SetMathAlphabet{\mathsfit}{bold}{\encodingdefault}{\sfdefault}{bx}{n}

\usepackage{url}
\usepackage{booktabs}
\usepackage{multirow}
\usepackage{ dsfont }
\usepackage{float}
\usepackage{color, colortbl, xcolor}
\usepackage{soul}
\usepackage{url}
\usepackage{multirow}
\usepackage{algorithm}
\usepackage{algorithmicx}
\usepackage{algpseudocode}
\usepackage{mathtools}
\usepackage{etoolbox}
\usepackage{lipsum}
\usepackage{bbm}
\usepackage{amssymb}
\usepackage{amsbsy}
\usepackage{pifont} 
\usepackage[normalem]{ulem}

\definecolor{2nd}{gray}{0.8}

\usepackage{nicefrac}       
\usepackage{lipsum}
\usepackage{fancyhdr}       
\usepackage{dirtytalk}
\usepackage{physics}
\usepackage{natbib}
\usepackage{times}
\usepackage{epsfig}
\usepackage{enumitem}
\usepackage[pagebackref,
breaklinks,
colorlinks=true,
citecolor=PineGreen,
linkcolor=BrickRed
]{hyperref}
\usepackage{pgfplots}
\usepackage{tikz}
\usetikzlibrary{
    positioning,
    shapes.geometric,
    arrows.meta
}
\usepackage{amsmath}
\usepackage{amssymb}
\usepackage{algpseudocodex}
\pgfplotsset{compat=newest}

\usepackage[dvipsnames]{xcolor}
\usepgfplotslibrary{fillbetween}
\usepackage{pgfplotstable}
\usepackage{hyperref}
\usepackage{amsmath}
\usepackage{amssymb}
\usepackage{mathtools}
\usepackage{amsthm}

\usepackage[capitalize]{cleveref}
\theoremstyle{plain}
\newtheorem{theorem}{Theorem}[section]

\theoremstyle{definition}

\theoremstyle{remark}
\newtheorem{remark}[theorem]{Remark}
\usepackage[textsize=tiny]{todonotes}

\title{Fatigue-Aware Learning to Defer via Constrained Optimisation}

\author{
 Zheng Zhang \\
  Centre for Vision, Speech and Signal Processing \\
  University of Surrey\\
  Guildford, Surrey, UK \\
   \And
 Cuong C. Nguyen \\
  Centre for Vision, Speech and Signal Processing \\
  University of Surrey\\
  Guildford, Surrey, UK \\
  \And
  David Rosewarne \\
  Centre for Vision, Speech and Signal Processing \\
  University of Surrey\\
  Guildford, Surrey, UK \\
  \And
 Kevin Wells \\
  Centre for Vision, Speech and Signal Processing \\
  University of Surrey\\
  Guildford, Surrey, UK \\
   \And
   Gustavo Carneiro\thanks{Corresponding authors} \\
   Centre for Vision, Speech and Signal Processing \\
  University of Surrey\\
  Guildford, Surrey, UK \\
}

\begin{document}
\maketitle

\begin{abstract}
Learning to defer (L2D) enables human–AI cooperation by deciding when an AI system should act autonomously or defer to a human expert. Existing L2D methods, however, assume static human performance, contradicting well‑established findings on fatigue‑induced degradation. We propose Fatigue‑Aware Learning to Defer via Constrained Optimisation (FALCON), which explicitly models workload‑varying human performance using psychologically grounded fatigue curves. FALCON formulates L2D as a Constrained Markov Decision Process (CMDP) whose state includes both task features and cumulative human workload, and optimises accuracy under human‑AI cooperation budgets via PPO‑Lagrangian training. We further introduce FA‑L2D, a benchmark that systematically varies fatigue dynamics from near‑static to rapidly degrading regimes. Experiments across multiple datasets show that FALCON consistently outperforms state‑of‑the‑art L2D methods across coverage levels, generalises zero‑shot to unseen experts with different fatigue patterns, and demonstrates the advantage of adaptive human–AI collaboration over AI‑only or human‑only decision‑making when coverage lies strictly between 0 and 1. Code is available at \href{https://github.com/zhengzhang37/FALCON.git}{https://github.com/zhengzhang37/FALCON.git}
\end{abstract}.
\section{Introduction}

AI systems are increasingly deployed in safety-critical applications, but relying solely on AI can be dangerous because they may overlook subtle issues that only humans can interpret. In domains such as financial risk assessment~\citep{green2019disparate}, breast cancer classification~\citep{halling2020optimam}, and detecting deceptive AI-generated content~\citep{ding2024hybrid}, human experts provide essential judgment and contextual understanding that current AI models cannot replicate. While AI offers consistent and relatively reliable performance, it can still make catastrophic errors that humans are better positioned to detect. Conversely, humans can be highly trustworthy in complex scenarios, but their performance is unstable and influenced by factors such as expertise level and fatigue.

Learning to defer (L2D) addresses these challenges by enabling \emph{hybrid intelligence} systems that dynamically allocate decisions between AI and human experts~\citep{fugener2022cognitive}. L2D methods learn a gating mechanism that defers high‑uncertainty cases to humans to maximise accuracy, while assigning high‑confidence cases to AI to reduce cost and conserve human effort~\citep{madras2018predict}. Existing L2D approaches are commonly categorised into \emph{one‑stage} and \emph{two‑stage} architectures: one‑stage methods jointly learn classification and deferral using shared representations~\citep{consistentest_Mozannar2020}, whereas two‑stage methods model these components separately~\citep{madras2018predict}. More recent work extends L2D to \emph{human‑adaptive} settings, enabling collaboration with previously unseen experts. For example, L2D‑Pop~\citep{tailor2024learning} conditions deferral on few‑shot \emph{context sets} of expert annotations, while EA‑L2D~\citep{strong2025expert} further simplifies this by representing experts via class‑level expertise estimated from the context set.

Despite these successes, most existing L2D methods rely on an unrealistic assumption: they treat \emph{human experts as static oracles with constant performance}. This simplification eases modelling but conflicts with extensive cognitive psychology evidence showing that human performance is dynamic, shaped by skill acquisition and, more critically, cognitive fatigue~\citep{casali2019rise,pimenta2014analysis,bose2019regression}. As task engagement accumulates, vigilance declines and accuracy degrades, a phenomenon known as the vigilance decrement~\citep{gyles2023psychometric}. Such fatigue‑induced performance degradation is well documented, particularly in prolonged or repetitive tasks~\citep{cairns2008double,lee2013cognitive}, and has serious real‑world consequences in domains such as radiology~\citep{reiner2012insidious,waite2017tired,taylor2019fatigue}. 
Berlin~et~al.~\citep{berlin2000liability}  reported a radiologist who made a critical misdiagnosis after interpreting 162 cases in a single day, which is more than triple the typical daily workload of 50 cases. Nevertheless, current L2D systems apply fixed deferral thresholds throughout a session, ignoring temporal variations in human reliability and potentially allocating equally difficult tasks to experts when they are fresh or fatigued without distinction.

\begin{figure}
    \centering
    \includegraphics[width=\linewidth]{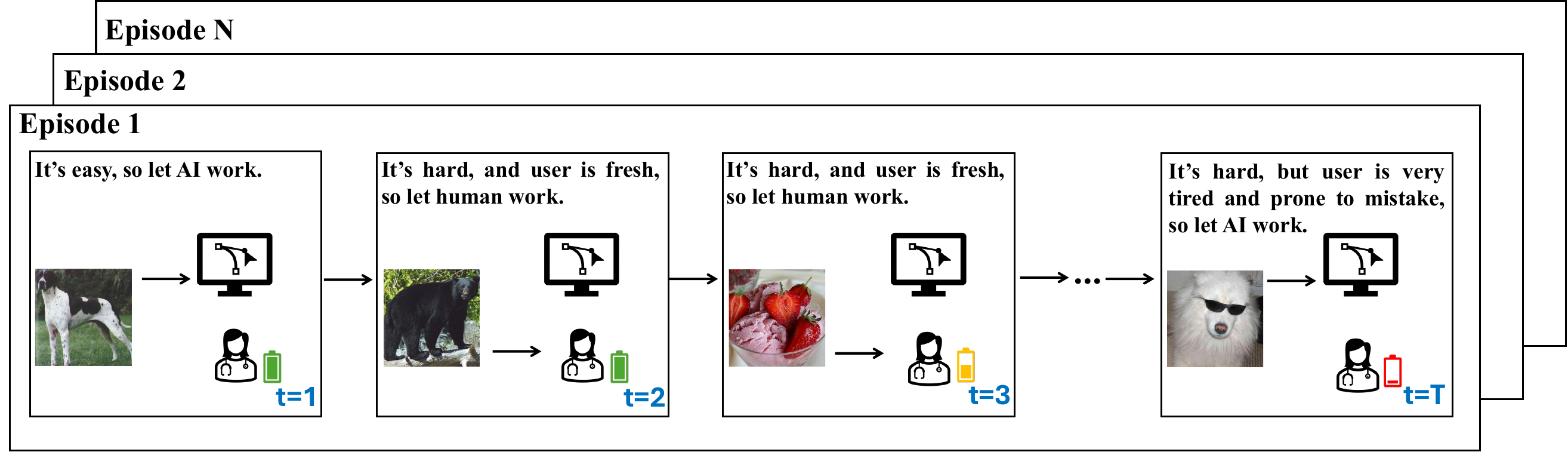}
    \vspace{-1ex}
    \caption{Example of an L2D scenario illustrating {workload-variant} human performance in human–AI task allocation within a single episode. FALCON adapts deferral decisions based on both task difficulty and accumulated human fatigue. At \(t=1\), an easy  task is handled by the AI while the human expert remains fresh. At \(t=2\), a challenging case is deferred to the human expert who has sufficient cognitive capacity. By \(t=3\), another hard task is still assigned to the human despite mild fatigue accumulation. At the final time step \(t=T\), severe human fatigue leads to AI handling the task to prevent performance degradation.}
    \label{fig:workflow}
    \vspace{-1ex}
\end{figure}

Inspired by cognitive psychology research on mental fatigue~\citep{estes2015workload,newell2013mechanisms}, we introduce a dynamic L2D setting, illustrated in \cref{fig:workflow}, that accounts for predictable variations in human performance, challenging the common assumption of static expert capability. We explicitly model {workload-dependent} human performance by linking expert accuracy to dynamic performance curves that capture both initial learning and subsequent fatigue-induced decline. To operationalise this, we introduce Fatigue-Aware Learning to Defer via Constrained Optimisation (FALCON), which formulates dynamic L2D as a Constrained Markov Decision Process (CMDP), where system states incorporate task-specific characteristics and cumulative human workload. This formulation enables our framework to make adaptive deferral decisions that align task allocation with the expert’s current cognitive state, rather than assuming static capability under a predetermined human-AI collaboration budget. Our main contributions can be summarised as follows:
\begin{itemize}[topsep=0pt, itemsep=0pt, leftmargin=3ex]
    \item \textbf{L2D with {workload-variant} human performance:} 
    FALCON is, to our knowledge, the first L2D framework that explicitly models workload-dependent human performance and requires a sequential CMDP formulation because each deferral decision changes the future state via an workload accumulator. This makes the deferral policy inherently stateful, where allocations alter subsequent human’s cognitive state, unlike prior L2D works that assume static human accuracy and thus non-sequential gating.    
    \item \textbf{Psychologically Grounded Simulation Environment:}
    We develop a human performance simulation environment grounded in psychological principles, offering a realistic testbed for evaluating L2D methods under {workload-variant} human performance conditions.
    \item \textbf{Fatigue-Aware L2D (FA-L2D) Benchmark:}
    We release the FA-L2D benchmark, based on Cifar100~\citep{wei2021learning}, Flickr~\citep{yang2017learning}, MiceBone~\citep{schmarje2022data}, and Chaoyang~\citep{zhu2021hard}, which models controllable fatigue effects across varying time horizons, enabling scenarios from near-constant to highly variable human performance and replacing prior benchmarks that assumed static human performance.
\end{itemize}
We evaluate FALCON against state-of-the-art L2D approaches~\citep{consistentest_Mozannar2020,madras2018predict,tailor2024learning,strong2025expert} on our proposed FA-L2D benchmark. Empirical results demonstrate that FALCON consistently outperforms existing methods, achieving higher accuracy for equivalent coverage levels across all evaluation settings. 
Importantly, under the {workload-variant} human performance proposed by our FA-L2D benchmark, L2D methods consistently outperform both AI-only and human-only decision-making for any non-trivial coverage level (i.e., between 0 and 1), highlighting the practical value of adaptive collaboration strategies.
\section{Preliminaries}

\subsection{Learning to Defer} 

For a \(K\)-way classification task, let \(\mathcal{D}=\{(\mathbf{x}_i, \mathbf{y}_i\})^{N}_{i=1}\) be the training set of size \(N\), where \(\mathbf{x}_i \in \mathcal{X} \subset \mathbb{R}^{d}\) denotes a \(d\)-dimensional input sample, and \(\mathbf{y}_i \in \mathcal{Y} \subset \{0, 1\}^{K}\) is the corresponding ground truth label. An \emph{AI classifier} is denoted as $\mathsf{m}:\mathcal{X} \to \Delta^{K-1}$, where a human expert is represented by \(\mathsf{h}:\mathcal{X} \to \Delta^{K-1}\).
Traditional L2D methods contain the classifier \(\mathsf{m}(\cdot)\) and a gating function \(\mathsf{g}(\cdot)\). 
Given an input sample \(\mathbf{x}\) and corresponding human prediction \(\mathsf{h}(\mathbf{x})\) and ground truth label \(\mathbf{y}\), the training objective is:
\begin{equation}
    \begin{aligned}[b]
        \ell(\mathsf{m},\mathsf{g}) = \mathbb{E}_{\mathbf{x},\mathbf{y},\mathbf{h}} \left[ (1-\mathsf{g}(\mathbf{x})) \mathbb{I}[\mathsf{h}(\mathbf{x}) \neq \mathbf{y}] \right. \left. + \mathsf{g}(\mathbf{x}) \mathbb{I} [ \mathsf{m}(\mathbf{x}) \neq \mathbf{y} ] \right],
    \end{aligned}
\label{eq:loss_1stage_L2D}
\end{equation}

where \(\mathbb{I}[\cdot]\) is the indicator function, \(\mathsf{g}(\mathbf{x}) \in [0,1]\) represents the probability that the AI classifier makes the prediction, while \(1-\mathsf{g}(\mathbf{x})\) denotes the probability deferring the decision to the human. Since \(\mathbb{I}[\cdot]\) is non-differentiable, some surrogate losses are proposed to generalise the cross-entropy loss~\citep{verma2022calibrated,mozannar2020consistent}.


Critically, all existing L2D methods are built on the simplifying assumption that the performance of the human prediction \(\mathsf{h}(\mathbf{x})\) is static over time, which is an assumption that ignores well-documented variations such as fatigue-induced degradation or learning effects~\citep{estes2015workload,leppink2019mental}, and thus fails to reflect realistic deployment conditions.

\subsection{Markov Decision Process} 
\label{sec:MDP}
A Markov Decision Process (MDP) can be described by a 4-tuple \((\mathcal{S}, \mathcal{A}, \mathsf{p}, \mathsf{r})\), where \(\mathcal{S}\) is the set of states called the \emph{state space}, \(\mathcal{A}\) is the set of actions called \emph{action space}, \(\mathsf{p}: \mathcal{S} \times \mathcal{A} \to \Delta(\mathcal{S})\) is the \emph{transition dynamics} with \(\Delta(\mathcal{S})\) being the probability simplex over \(\mathcal{S}\), and \(\mathsf{r}: \mathcal{S} \times \mathcal{A} \times \mathcal{S} \to \mathbb{R}\) is a \emph{reward function}. A \emph{policy} \(\pi : \mathcal{S} \to \Delta(\mathcal{A})\) maps a state in \(\mathcal{S}\) to a probability distribution over the actions in \(\mathcal{A}\). An \emph{optimal policy} \(\pi^*\) is a policy that maximises the expected value of the discounted return \(J_{\mathsf{r}}(\pi) = \mathbb{E}_{\mathbf{s}_{0} \sim \mathcal{S}} [\sum_{t = 0}^{\infty} \gamma^{t} \mathsf{r}(\mathbf{s}_t, \pi(\mathbf{s}_t), \mathbf{s}_{t + 1} )]\), where \(\gamma \in [0, 1]\) is a discount factor. The value function is defined as \(V^\pi_{\mathsf{r}}(\mathbf{s}) = \mathbb{E}_{\tau\sim\pi}[\sum_t\gamma^t \mathsf{r}(\mathbf{s}_t, \mathbf{a}_t, \mathbf{s}_{t+1})|\mathbf{s}_0=\mathbf{s}]\), the action-value function is defined as \(Q^\pi_{\mathsf{r}}(\mathbf{s}, \mathbf{a}) = \mathbb{E}_{\tau\sim\pi}[\sum_t\gamma^t \mathsf{r}(\mathbf{s}_t, \mathbf{a}_t, \mathbf{s}_{t+1})|\mathbf{s}_0=\mathbf{s}, \mathbf{a}_0=\mathbf{a}]\) and the advantage function is defined as \(A^\pi_{\mathsf{r}}(\mathbf{s}, \mathbf{a}) = Q^\pi_{\mathsf{r}}(\mathbf{s}, \mathbf{a}) - V^\pi_{\mathsf{r}}(\mathbf{s})\).

Constrained Markov decision process (CMDP) is an augmented version of MDP~\citep{altman2021constrained}, 
defined by the tuple \((\mathcal{S,A,C}, \mathsf{p}, \mathsf{r})\), in which the set of constraints is defined as:
\( \mathcal{C} = \left\{ \pi \in \Pi \Big| J_{\mathsf{c}_i}(\pi) \le d_i, i \in \{1,\dots,C\}   \right\} \), where 
\(J_{\mathsf{c}_i}(\pi) = \mathbb{E}_{\tau \sim \pi} \left[ \sum_{t} \gamma_{\mathsf{c}_{i}}^{t} \mathsf{c}_{i}(\mathbf{s}_t,\mathbf{a}_t) \right]\), with \(\mathsf{c}_{i}:\mathcal{S} \times \mathcal{A} \times \mathcal{S} \to \mathbb{R}\).
The training objective is then defined as \(\max_{\pi \in \mathcal{C}}J_{\mathsf{r}}(\pi)\), where \(\mathcal{C}\) is the constraint (or feasible) set. In this setting, the corresponding value function, action-value function, and advantage functions for the auxiliary
costs are denoted by \(V^\pi_{\mathsf{c}}(\mathbf{s})\), \(Q^\pi_{\mathsf{c}}(\mathbf{s}, \mathbf{a})\), \(A^\pi_{\mathsf{c}}(\mathbf{s}, \mathbf{a})\).

\section{Methodology}

In this section, we present FALCON, a framework that formulates L2D as a CMDP to address the human-AI cooperation with human performance degradation dependent on workload accumulation. Firstly, we define the human-AI collaborative sequential decision-making task. We then introduce a human performance simulation environment grounded in psychological principles. Lastly, we illustrate the L2D architecture with {workload-variant} human performance, while introducing constrained optimisation for precise budget control over human-AI cooperation costs.

\subsection{Environment Setup}
\label{sec:task_definition}

We address sequential classification in the form of episodes. In each episode, a human-AI team collaboratively processes a stream of \(T\) sequential data \(\tau = {\{(\mathbf{x}_t, \mathbf{y}_t)\}}_{t=1}^{T}\)
, where \(\mathbf{x}_t \in \mathcal{X} \subset \mathbb{R}^{d}\) is an input sample at time step \(t\), and \(\mathbf{y}_t \in \mathcal{Y} = \{1, \dots, K\}\) is the corresponding ground truth label.
The system maintains two predictive components: 
1) a human expert with workload affected performance by  defined by \(\mathsf{h}:\mathcal{X} \times \mathcal{W} \to \mathcal{Y} \), where \(\mathcal{W} \subset \mathbb{R}_{+}\) is the space that represents the cumulative human workload, and 2) the AI classifier defined by \(\mathsf{m}:\mathcal{X} \to \Delta^{K-1}\).  At each time step \(t\), the system will perform an action \(\mathbf{a}_t \in \{\text{AI}, \text{Human}\}\). This action determines which agent will produce the final prediction \(\hat{\mathbf{y}}_t\) of the sample \(\mathbf{x}_t\).

\subsection{Human Performance Simulation } 
\label{sec:human_performance}

The human performance is simulated with two key assumptions: (1) \emph{Predictable Fatigue Accumulation}~\citep{estes2015workload}, where human cognitive performance degrades as a function of cumulative engagement in decision-making tasks, following psychologically grounded fatigue curves; and (2) \emph{Selective Fatigue}~\citep{hopko2021effect}, where only tasks assigned to the human expert contribute to fatigue accumulation, while tasks handled by the AI system impose no additional cognitive load. 

\begin{figure}
    \centering
    \vspace{-1em}
    \begin{subfigure}[b]{0.48\linewidth}
        \centering
        \includegraphics[width=\linewidth]{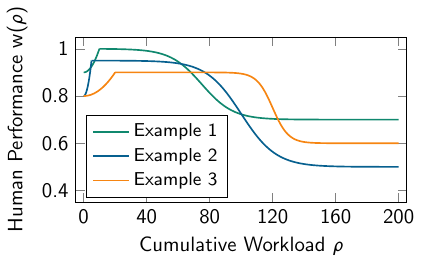}
        \vspace{-2em}
        \caption{}
        \label{fig:fatigue_curves}
    \end{subfigure}
    \begin{subfigure}[b]{0.48\linewidth}
    \centering
    \resizebox{\textwidth}{!}{
        \includegraphics[width=\linewidth]{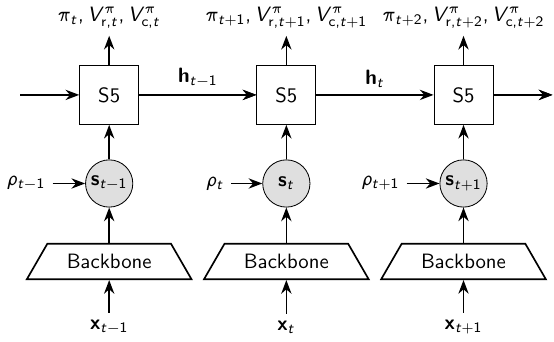}}
        \vspace{-2em}
        \caption{}
        \label{fig:architecture}
    \end{subfigure}
    \vspace{-1ex}
    \caption{(a): Examples of \(\mathsf{w}(\rho)\). The values of parameters \((w_{0}, w_\text{peak}, w_\text{base}, k, \bar{\rho}, \hat\rho)\) in Example 1,2 and 3 are \((0.9, 1, 0.7, 0.1, 0.375, 0.05)\), \((0.8, 0.95, 0.5, 0.09, 0.5, 0.025)\) and \((0.8, 0.9, 0.6, 0.2, 0.6, 0.1)\). (b): The architecture of FALCON with {workload-variant} human performance. A backbone model extracts visual features from the input \(\mathbf{x}_t\), while the cumulative human workload \(\rho_t\) is passed through an embedding layer. The visual and workload features are concatenated and processed by a Resettable S5 layers~\citep{lu2023structured} to capture temporal dependencies and output the policy \(\pi(\mathbf{a}_t|\mathbf{s}_t)\) alongside value estimates. 
    }
    \vspace{-1ex}
\end{figure}

\textbf{Mental Fatigue Curves} \quad Since vigilance wanes as cognitive fatigue accumulates~\citep{mccarley2021psychometric,gyles2023psychometric}, we model human performance \(\mathsf{w}: \mathcal{W} \to [0,1]\) using a two-phase piece-wise function:
\begin{equation}
    \scalebox{1.}{$
    \mathsf{w}(\rho) = \begin{cases} 
            w_{0} + (w_{\text{peak}} - w_{0}) \left(\frac{\rho}{\hat{\rho} \cdot L}\right)^2 & \mbox{if } 0 \le\rho \le \hat \rho \cdot L\\
            w_{\text{base}} + (w_{\text{peak}} - w_{\text{base}}) \frac{1}{1 + \exp[ k(\rho-\bar{\rho}\cdot L) ]} & \mbox{if }   \rho \ge \hat \rho \cdot L\\
          \end{cases},$}
    \label{eq:W_H}
\end{equation}
where \(w_{0}, w_{\text{peak}}, w_{\text{base}}\) denote the initial, peak and minimum (or base) performance levels, \(\rho \in \mathcal{W}\) is the cumulative workload (see \cref{eq:cumulative_workload}), and \(\hat \rho, \bar{\rho}\) denote the relative workload at the peak performance and at the inflection point of the decay phase, and \(k\) is the steepness of performance decline.
The two‑phase warm‑up + sigmoid fatigue curve, represented by \(\mathsf{w}(\rho)\) in \cref{eq:W_H}, models how human performance evolves with cumulative workload \(\rho\) over \(L\) time steps. Grounded in cognitive psychology, it captures an initial warm‑up phase, modelled as quadratic growth~\citep{newell2013mechanisms}, where performance improves from \(w_0\) to peak \(w_{\text{peak}}\), followed by a fatigue phase, modelled as sigmoid decay~\citep{estes2015workload}, where performance degrades toward \(w_{\text{base}}\). Together, these phases provide a controlled yet principled setting for evaluating L2D policies under workload‑dependent human performance.
The parameters \((w_0, w_{\text{peak}}, w_{\text{base}}, \hat{\rho}, \bar{\rho}, k)\) can be \emph{robustly} specified in practice from established cognitive psychology literature~\citep{newell2013mechanisms,estes2015workload}. Crucially, \textsc{FALCON} does \emph{not} require user-specific parameter fitting at deployment since we learn policies that are robust to parameter uncertainty, supporting \emph{zero-shot generalisation} to unseen experts and fatigue patterns, as demonstrated in \cref{sec:experiments}.
Three different human performance curves are shown in \cref{fig:fatigue_curves} by varying the parameters in \cref{eq:W_H}.

\textbf{Human Prediction Modelling} \quad
Given the human performance at a particular workload at time step \(t\), defined as \(\mathsf{w}(\rho_t)\), we model human prediction errors via noise rate \(\eta\), which represents the probability at time step \(t\) of a classification error, defined as \(\eta_t = 1 - \mathsf{w}(\rho_t)\). The prediction distribution of the  human prediction given noise rate \(\eta_t\) is defined as:
\begin{equation}
    \Pr(\hat{\mathbf{y}}|\mathbf{y},\mathbf{x},\eta_t)=(1-\eta_t)\cdot\mathbb{I}(\hat{\mathbf{y}}=\mathbf{y})+\nicefrac{\eta_t}{K-1}\cdot\mathbb{I}(\hat{\mathbf{y}}\neq\mathbf{y}),
\label{eq:sample_human_prediction}
\end{equation}
where \(\mathbf{x}\) and \(\mathbf{y}\) denote the data sample and ground truth label, respectively. This means that the human predicts the ground truth label with probability \( (1-\eta_t) \), and one of the \(K-1\) incorrect labels with probability \( \nicefrac{\eta_t}{K-1} \).

\textbf{Scope of realism and recovery} \quad Our model is a first step toward realism: it encodes warm up and non linear fatigue within a session, consistent with established findings~\cite{newell2013mechanisms,estes2015workload}, while recovery is negligible within a session and reset across episodes to approximate overnight recovery (episodic setup in \cref{sec:task_definition}). This choice isolates decision dependent fatigue effects on deferral policy learning without introducing additional confounds from short term recovery dynamics.

\subsection{Fatigue-Aware
Learning to Defer via Constrained Optimisation}

We model this workflow as a CMDP, where the state space is \(\mathbf{s}_t=(\mathbf{x}_t,\rho_t)\in\mathcal{X}\times\mathcal{W}\), where \(\mathbf{x}_t\) and \(\rho_t\) denote the current input sample and cumulative human workload, respectively. The system transitions deterministically based on the workload update rule: 
\begin{equation}
\rho_{t+1} = \begin{cases}
\rho_t + 1 & \text{if } \mathbf{a}_t = \text{Human} \\ \
\rho_t & \text{if } \mathbf{a}_t = \text{AI}
\end{cases},    
\label{eq:cumulative_workload}
\end{equation}
where \(\rho_1 = 0\). 
The reward function is denoted by prediction accuracy (i.e., \(\mathsf{r}(\mathbf{s}_t, \mathbf{a}_t)=\mathbb{I}[\hat{\mathbf{y}}_t=\mathbf{y}_t]\)), where \(\hat{\mathbf{y}}_t\) is the final decision of the system at the time step \(t\), while the constraint set \(\mathcal{C}\) defines lower and upper limits to human workload, 
where the lower bound is denoted by \( \sum_{t=1}^T \mathsf{c}(\mathbf{s}_t, \mathbf{a}_t) \ge d_l \) and the upper limit is defined by \( \sum_{t=1}^T \mathsf{c}(\mathbf{s}_t, \mathbf{a}_t) \le d_u \), with \(\mathsf{c}(\mathbf{s}_t, \mathbf{a}_t)=\mathbb{I}[\mathbf{a}_t=\text{Human}]\).

\textbf{L2D Architecture with {workload-variant} Human Performance} The architecture of our L2D architecture with {workload-variant} human performance (\cref{fig:architecture}) employs an actor-critic strategy for adaptive decisions. A backbone model takes the input sample \(\mathbf{x}_t\) to extract visual feature embeddings, while the cumulative workload \(\rho_t\) is embedded by a learnable  linear layer. Then the visual and workload features are concatenated and processed through Resettable simplified structured state space sequence (S5) layers~\citep{lu2023structured}, which represent a variation of structured state space sequence (S4) models~\citep{smith2023simplified, gu2022efficiently}, to capture temporal dependencies and maintain memory of the human's cognitive state trajectory--this is represented by the state vector \(\mathbf{h} \in \mathcal{H}_t \subset \mathbb{R}^{H}\).
From this state vector, three distinct heads predict: the policy \(\pi_t=\pi(\mathbf{a}_t|\mathbf{s}_t)\), the estimated future reward \({V}^{\pi}_{\mathsf{r}, t}={V}^{\pi}_{\mathsf{r}}(\mathbf{s}_t)\), and the estimated future cost \({V}^{\pi}_{\mathsf{c}, t}={V}^{\pi}_{\mathsf{c}}(\mathbf{s}_t)\).

\textbf{Constrained Optimisation with PPO-Lagrangian} \quad
We formulate the training phase as a constrained optimisation problem:
\begin{equation}
    \textstyle \max_{\pi_\theta \in \Pi}J_{\mathsf{r}}(\pi_{\boldsymbol{\theta}}) \qquad \text{s.t.} \quad d_{l} \leq J_{\mathsf{c}}(\pi_{\boldsymbol{\theta}}) \leq d_{u},
    \label{eq:objective}
\end{equation}
where \(J_{\mathsf{r}}(\pi_{\boldsymbol{\theta}})\) is defined in \cref{sec:MDP}, \(J_{\mathsf{c}}(\pi_{\boldsymbol{\theta}}) = \mathbb{E}_{\tau\sim\pi_{\boldsymbol{\theta}}}[\sum_t\gamma^t \mathsf{c}(\mathbf{s}_t, \mathbf{a}_t)]\), with \(\mathsf{c}(.)\) defined in \eqref{eq:cumulative_workload}, and \(d_l,d_u\) represent the lower and upper limits in cumulated workload.
Following the PPO-Lagrangian method~\citep{ray2019benchmarking}, the constrained problem in \cref{eq:objective} can be solved via the Lagrangian dual formulation~\citep{altman1998constrained}:
\begin{equation}
\scalebox{1.}{$
\begin{aligned}
    \textstyle \min_{\lambda_u,\lambda_l \geq 0}\max_{\pi_{\boldsymbol{\theta}} \in \Pi}J_{\mathsf{r}}(\pi_{\boldsymbol{\theta}}) &- \lambda_u \cdot \max(0, J_{\mathsf{c}} (\pi_{\boldsymbol{\theta}}) - d_u) \\
    & -  \lambda_l \cdot \max(0, -J_{\mathsf{c}}(\pi_{\boldsymbol{\theta}}) + d_l).
    \label{eq:optimisation_dual_lagrangian}
\end{aligned}$}
\end{equation}

The optimisation of \eqref{eq:optimisation_dual_lagrangian} involves the update of the Lagrangian multipliers with gradient ascent. This formulation explicitly enforces lower and upper workload bounds \((d_{l},d_{u})\), enabling precise control of human utilisation (coverage, which is equal to \(1 - (\sum_{t=1}^T \mathsf{c}(\mathbf{s}_t, \mathbf{a}_t))/T\))) without ad‑hoc threshold tuning. Note that \(\gamma_{\mathsf{c}_{i}}^{t}\) is set to 1 to align the constraint with the intended coverage.
If the agent defers too much (exceeding the \(d_{u}\)), \(\lambda_u\) increases, which heavily penalises the deferral action in the loss function. 
If the agent defers too little (below the \(d_{l}\)), \(\lambda_l\) increases, encouraging  deferral. 
Because the objective is to maintain the expected cost within the budget interval, the same cost term \(J_{\mathsf{c}}(\pi_{\boldsymbol{\theta}})\) is shared by both multipliers. This mechanism yields a target coverage (e.g., 40\%) by automatically penalising over‑ and under‑deferral during training; for example, we set \(d_{u}=0.65\) and \(d_{l}=0.55\) to obtain approximately 0.4 coverage.

Specifically, the critic model is optimised by regression on mean-square error between value estimator and the true trajectory value, followed by the standard PPO setting. The policy update follows standard PPO-Lagrangian with modified objective:
\begin{align}
    J_{\mathsf{r}}(\pi_{\boldsymbol{\theta}}) = \mathbb{E}_{\tau \sim \pi_{\boldsymbol{\theta}_{old}}} \left[\min \left ( \frac{\pi_{\boldsymbol{\theta}}(\mathbf{a}|\mathbf{s})}{\pi_{\boldsymbol{\theta}_{old}}(\mathbf{a}|\mathbf{s})}A_{\mathsf{r}}^{\pi_{\boldsymbol{\theta}_{old}}}(\mathbf{s}, \mathbf{a}), \text{clip} \left ( \frac{\pi_{\boldsymbol{\theta}}(\mathbf{a}|\mathbf{s})}{\pi_{\boldsymbol{\theta}_{old}}(\mathbf{a}|\mathbf{s})}, 1-\epsilon, 1+\epsilon\right ) A_{\mathsf{r}}^{\pi_{\boldsymbol{\theta}_{old}}}(\mathbf{s}, \mathbf{a})\right )\right ] \\
    J_{\mathsf{c}}(\pi_{\boldsymbol{\theta}}) = \mathbb{E}_{\tau \sim \pi_{\boldsymbol{\theta}_{old}}} \left[\min \left ( \frac{\pi_{\boldsymbol{\theta}}(\mathbf{a}|\mathbf{s})}{\pi_{\boldsymbol{\theta}_{old}}(\mathbf{a}|\mathbf{s})}A_{\mathsf{c}}^{\pi_{\boldsymbol{\theta}_{old}}}(\mathbf{s}, \mathbf{a}), \text{clip} \left ( \frac{\pi_{\boldsymbol{\theta}}(\mathbf{a}|\mathbf{s})}{\pi_{\boldsymbol{\theta}_{old}}(\mathbf{a}|\mathbf{s})}, 1-\epsilon, 1+\epsilon\right ) A_{\mathsf{c}}^{\pi_{\boldsymbol{\theta}_{old}}}(\mathbf{s}, \mathbf{a})\right )\right ]
    \label{eq:policy_update}
\end{align}

\noindent
where \(\boldsymbol{\theta}_{old}\) is the vector of policy parameters before the update, \(\text{clip}(.)\) is the clipping operation on the probability ratio with clipping parameter \(\epsilon\), which controls the maximum allowed deviation of the updated policy from the previous policy to ensure stable training.

Furthermore, we update the Lagrangian multipliers using gradient ascent with an Adam optimiser:

\begin{align}
    \lambda_u^{(\beta+1)} &= \max(0, \lambda_u^{(\beta)} + \alpha_\lambda \cdot ( J_\mathsf{c}(\pi) - d_u)) \\
    \lambda_l^{(\beta+1)} &= \max(0, \lambda_l^{(\beta)} + \alpha_\lambda \cdot ( d_l - J_\mathsf{c}(\pi)))
    \label{eq:Lagrangian}
\end{align}

where \(\lambda_u\) and \(\lambda_l\) are the Lagrangian multipliers for the upper and lower bound constraints respectively, \(\alpha_\lambda\) is the learning rate for the multiplier updates, $d_u$ and \(d_l\) denote the upper and lower constraint thresholds, \(\beta\) denotes the update step, and \(J_{\mathsf{c}}(\pi)\) represents the expected cumulative cost under policy \(\pi\). The multipliers automatically adjust to enforce the constraint bounds: \(\lambda_u\) increases when the cost exceeds the upper limit \(d_u\), penalizing excessive human utilisation, while \(\lambda_l\) increases when the cost falls below the lower limit \(d_l\), encouraging sufficient human engagement.

\textbf{Design Choice of S5 for FALCON} \quad
Modelling human cognitive state over long episodes is critical for FALCON to track cumulative fatigue, making the choice of sequence model critical. RL systems typically employ RNNs~\citep{yan2023efficient,jha2025cross,gessler2025overcookedv2,morad2023popgym,david2022decision,lu2023structured}, Transformers~\citep{chen2021decision,parisotto2020stabilizing}, and as the sequence models. However, traditional RNNs, such as LSTM and GRU, suffer vanishing gradients over long sequences while Transformers incur quadratic computational costs prohibitive for extended episodes~\citep{metz2021gradients}
We employ Resettable Simplified Structured State Space Sequence (S5) layers~\citep{lu2023structured}, a variant of S4 models~\citep{smith2023simplified, gu2022efficiently}, which provide linear computational complexity and stable gradient flow essential for tracking cumulative workload over hundreds of time steps. S5 demonstrates superior asymptotic runtime compared to Transformers while significantly outperforming LSTMs in both performance and computational efficiency~\citep{lu2023structured}. 


\begin{algorithm*}
\caption{FALCON training procedure}
\label{alg:human_ai_collaboration_training}
    \begin{algorithmic}[1]
        \Procedure{Training}{$\mathcal{D}, n_{\text{iter}}, n_{\text{episode}}$}
            \LComment{\(\mathcal{D}={\{\mathbf{x}_t, \mathbf{y}_t\}}_{t=1}^{T}\): training dataset}
            \LComment{\(n_{\text{iter}}\): the total number of iterations}
            \LComment{\(n_{\text{episode}}\): number of episodes}
            \State initialise AI classifier \(\mathsf{m}\), policy \(\pi_{\boldsymbol{\theta}_1}\),value function \(V_\mathsf{r}^{\boldsymbol{\phi}_1}\) and cost value function \(V_\mathsf{c}^{\boldsymbol{\psi}_1}\)
            \State initialise Lagrangian multiplier \(\lambda_u\), \(\lambda_l\)

            \For{\(j = 1\) to \(n_{\text{iter}}\)}
                \State collect set of trajectories: \(\hat{\mathcal{D}}_j \gets \Call{Collect Trajectories}{\mathcal{D}, \mathsf{m}, \pi_{\boldsymbol{\theta}}, V_\mathsf{r}^{\boldsymbol{\phi}}, V_\mathsf{c}^{\boldsymbol{\psi}}, n_{\text{episode}}}\)
            \State update Lagrangian multiplier \(\lambda_u\), \(\lambda_l\) via gradient ascent \Comment{defined in \cref{eq:Lagrangian}}
            \State compute estimated reward value \(\hat{\mathsf{r}}_t=\sum_{j=0}^{T-t}\gamma^j\mathsf{r}_{t+j}\) and reward advantage \(A_{\mathsf{r}}^{\pi_{\boldsymbol{\theta}_i}}\)
            \State compute estimated cost value \(\hat{\mathsf{c}}_t=\sum_{j=0}^{T-t}\gamma^j\mathsf{c}_{t+j}\) and cost advantage \(A_{\mathsf{c}}^{\pi_{\boldsymbol{\theta}_j}}\)
            \State shuffle data in \(\hat{\mathcal{D}}_j\) and split into mini-batches
            \For{each mini-batch from \(\hat{\mathcal{D}}_j\)}
                \State update \(\pi_{\boldsymbol{\theta}_{j+1}}\) using PPO
                \Comment{defined in \cref{eq:policy_update}}
                \State update reward value function: \(\boldsymbol{\phi}_{j+1} \gets \arg \min_{\boldsymbol{\phi}} \frac{1}{|\hat{\mathcal{D}}_j|T} \sum (V_\mathsf{r}^{\boldsymbol{\phi}_j} - \hat{\mathsf{r}}_t)^2\)
                \State update cost value function: \(\boldsymbol{\psi}_{j+1} \gets \arg \min_{\boldsymbol{\psi}} \frac{1}{|\hat{\mathcal{D}}_j|T} \sum (V_\mathsf{c}^{\boldsymbol{\psi}_j} - \hat{\mathsf{c}}_t)^2\)
            \EndFor
        \EndFor
        \State \Return the optimal policy \(\boldsymbol{\theta}_{n_{iter}}\)
        \EndProcedure
        \Statex
        \Procedure{Collect Trajectories}{\(\mathcal{D}, \mathsf{m}, \pi_{\boldsymbol{\theta}}, V_\mathsf{r}^{\boldsymbol{\phi}}, V_\mathsf{c}^{\boldsymbol{\psi}}, n_{\text{episode}}\)}
            \LComment{\(\mathcal{D}\): training dataset}
            \LComment{\(\mathsf{m}\): AI classifier}
            \LComment{\(\pi_{\boldsymbol{\theta}}\): policy function parameterised by \(\boldsymbol{\theta}\)}
            \LComment{\(V_\mathsf{r}^{\boldsymbol{\phi}}\): reward value function parameterised by \(\boldsymbol{\phi}\)}
            \LComment{\(V_\mathsf{c}^{\boldsymbol{\psi}}\): cost value 
            function parameterised by \(\boldsymbol{\psi}\)}
            \LComment{\(n_{\text{episode}}\): number of episodes}
            \State set data buffer \(\hat{\mathcal{D}}=\varnothing\)
            \For{\(i = 1\) to \(n_{\text{episode}}\)}
                \State sample a sequences of \(T\) images from \(\mathcal{D}\)
                \State sample fatigue model parameters \(w_{0}, w_{peak}, w_{base}, k, \bar{\rho}, \hat\rho\)
                \Comment{See \cref{tab:human_params_cifar100,tab:human_params_chaoyang,tab:human_params_flickr,tab:human_params_micebone}}
                \State initialise human workload accumulator \(\rho \leftarrow 0\) and \(\mathsf{w}(0) \leftarrow w_{0}\)
                \For{\(t = 1\) to \(T\)}
                    \State get current state: \(\mathbf{s}_t \gets (\mathbf{x}_t, \rho\))
                    \State sample an action from the policy: \(\mathbf{a}_t \sim \pi_{\boldsymbol{\theta}_i}(\mathbf{s}_t)\)
                
                    \If{\(\mathbf{a}_t = \text{human}\)}
                        \Comment{human expert makes the prediction}
                        \State update human workload: \(\rho \leftarrow \rho + 1\) 
                        \State update human performance: \(\mathsf{w}_t \leftarrow \mathsf{w}(\rho)\) \Comment{defined in \cref{eq:W_H}}
                        \State get the annotation flipping probability of human due to fatigue: \(\eta \leftarrow 1-\mathsf{w}_t\)
                        \State sample human prediction: \(\hat{\mathbf{y}}_t \sim \text{Pr}(\hat{\mathbf{y}_t}|\mathbf{y}_t,\eta)\) \Comment{defined in \cref{eq:sample_human_prediction}} 
                    \ElsIf{\(\mathbf{a}_t = \text{AI}\)}
                        \Comment{AI classifier makes the prediction}
                        \State get the label predicted by the classifier: \(\hat{\mathbf{y}}_t \gets \operatorname*{argmax}_{} \mathsf{m}(\mathbf{x}_t) \)
                    \EndIf
                    
                    \State \(\mathsf{r}_t \gets \mathbb{I}(\mathbf{y}_t=\hat{\mathbf{y}}_t)\) 
                    \State gather data from \(\pi(\cdot|\mathbf{s}_t, \mathbf{a}_t)\), then \(\hat{\mathcal{D}}=\hat{\mathcal{D}}\cup\{\tau_{t+1}, \mathbf{s}_t, \mathbf{a}_t, \mathsf{r}_t, \log \pi_{\boldsymbol{\theta}}(\mathbf{a}_t | \mathbf{s}_t), V_r^{\boldsymbol{\phi}}, V_c^{\boldsymbol{\psi}}\}\)
                \EndFor
            \EndFor
            
            \State \Return \(\hat{\mathcal{D}}\)
        \EndProcedure
    \end{algorithmic}
\end{algorithm*}

\begin{algorithm*}
\caption{FALCON testing procedure}
\label{alg:human_ai_collaboration_testing}
    \begin{algorithmic}[1]
        \Procedure{Testing}{$\mathcal{D}, \mathsf{w}, \mathsf{m}, \boldsymbol{\theta}$}
            \LComment{\(\mathcal{D}={\{\mathbf{x}_t, \mathbf{y}_t\}}_{t=1}^{T}\): testing dataset}
            \LComment{\(\boldsymbol{\theta}\): parameter of policy function}
            \LComment{\(\mathsf{w}\): fatigue function}
            \LComment{\(\mathsf{m}\): AI classifier}
            \State sample fatigue model parameters \(w_{0}, w_{peak}, w_{base}, k, \bar{\rho}, \hat\rho\)
                \Comment{See \cref{tab:human_params_cifar100,tab:human_params_chaoyang,tab:human_params_flickr,tab:human_params_micebone}}
            \State initialise human workload accumulator \(\rho \leftarrow 0\) and \(\mathsf{w}(0) \leftarrow w_{0}\)
            \State initialise accumulate accuracy \(\mathsf{r}\)
            \For{\(t = 1\) to \(T\)}
                \State get current state: \(\mathbf{s}_t \gets (\mathbf{x}_t, \rho\))
                \State select an action: \(\mathbf{a}_t \gets\operatorname*{argmax}_{\boldsymbol{\theta}} \pi_{\boldsymbol{\theta}}(\mathbf{s}_t)\)
            
                \If{\(\mathbf{a}_t = \text{human}\)}
                    \Comment{human expert makes the prediction}
                    \State update human workload: \(\rho \leftarrow \rho + 1\) 
                    \State update human performance: \(\mathsf{w}_t \leftarrow \mathsf{w}(\rho)\) \Comment{defined in \cref{eq:W_H}}
                    \State get the annotation flipping probability of human due to fatigue: \(\eta \leftarrow 1-\mathsf{w}_t\)
                    \State sample human prediction: \(\hat{\mathbf{y}}_t \sim \Pr(\hat{\mathbf{y}}_t|\mathbf{y}_t,\eta)\) \Comment{defined in \cref{eq:sample_human_prediction}} 
                \ElsIf{\(\mathbf{a}_t = \text{AI}\)}
                    \Comment{AI classifier makes the prediction}
                    \State get the label predicted by the classifier: \(\hat{\mathbf{y}}_t \gets \operatorname*{argmax}_{} \mathsf{m}(\mathbf{x}_t) \)
                \EndIf
                \Statex
                \LComment{Calculate accuracy \(\mathsf{r}_t\)}
                \If{\(\hat{\mathbf{y}}_t = \mathbf{y}_t\)}
                    \State \(\mathsf{r}_t \leftarrow 1\) \Comment{correct prediction}
                \Else
                    \State \(\mathsf{r}_t \leftarrow 0\) \Comment{incorrect prediction}
                \EndIf
                \State \({\mathsf{r}} \gets \hat{\mathsf{r}} + \mathsf{r}_t\) \Comment{accumulate reward}
            \EndFor
        \State \Return \({\mathsf{r}/T}, 1-\rho/T\) \Comment{return accuracy and coverage}
        \EndProcedure
    \end{algorithmic}
\end{algorithm*}

\section{Fatigue-Aware L2D (FA-L2D) Benchmark}
\label{sec:FA_L2D_Benchmark}

\vspace{-1ex}
Our new benchmark is designed to evaluate L2D methods under the assumption that humans exhibit variable performance as a function of cumulative workload. 
During each training episode, images are randomly sampled from the training set, while human performance parameters are randomly sampled from predefined ranges (See \cref{tab:human_params_cifar100,tab:human_params_chaoyang,tab:human_params_flickr,tab:human_params_micebone} for dataset-specific human performance parameter ranges). As fatigue accumulates according to \cref{eq:W_H}, human predictions are modelled probabilistically with a noise rate \(\eta_t = 1 - \mathsf{w}(\rho_t)\), while humans predict correctly with probability \(1-\eta_t\) and make random classification errors among the remaining \(K-1\) classes with probability \(\nicefrac{\eta_t}{K-1}\).
This design systematically varies curve parameters to generate diverse scenarios spanning near‑static, normal, and rapid‑fatigue regimes, enabling stress tests of L2D methods under plausible human performance trajectories and providing a more realistic representation of human–AI cooperation environments than prior benchmarks that assume temporal stability.
The controllable nature of fatigue parameters allows systematic evaluation of L2D across different human performance profiles, from minimally fatigued experts to those experiencing significant cognitive decline over time.

\textbf{Datasets} \quad \emph{Cifar100}~\citep{krizhevsky2009learning}  has 50k training images and 10k testing images, with each image belonging to one of 100 classes.  \emph{Chaoyang}~\citep{zhu2021hard} comprises 6,160 colon slide patches categorised into four classes. 
\emph{MiceBone}~\citep{schmarje2022data} has 7,240 second-harmonic generation microscopy images, where the annotation consists of one of three possible classes. 
\emph{FLickr10K}~\citep{yang2017learning} is a large-scale dataset containing 10,700 images labelled with 8 commonly used emotions.
To ensure fair comparison across all methods, we standardise the testing episodes by reshuffling several datasets. Please refer to \cref{sec:data} for the datasets details.

\textbf{Metrics} \quad 
We evaluate performance using prediction accuracy as a function of coverage on test set episodes, where coverage denotes the percentage of samples classified solely by the AI. These accuracy–coverage curves capture the trade-off between accuracy and cooperation budget as coverage varies from 0\% (human-only classification) to 100\% (AI-only classification). Reported results are averaged over three models trained with different random seeds and evaluated at the final training epoch.
To provide a concise quantitative summary, we compute the \emph{area under the accuracy–coverage curve} (AUACC), where higher AUACC indicates more favourable accuracy–coverage trade-offs.

\textbf{Ablation Settings} \quad 
To systematically evaluate L2D methods across different human performance patterns, we define three distinct benchmark cases that capture varying degrees of fatigue-induced performance degradation as follows:
\begin{enumerate}[topsep=0pt, itemsep=0pt,leftmargin=5ex]
    \item \emph{Sustained High Performance:} This case models scenarios where human experts maintain consistently high performance throughout the task duration, approximating the static expert assumption used in traditional L2D methods.
    \item \emph{Normal Fatigue:} This case represents typical workplace conditions where human performance follows a standard warm-up and fatigue cycle. 
    \item \emph{Rapid Fatigue:} This case simulates sharp decline in human performance, aiming to test the robustness of L2D methods under extreme cognitive fatigue conditions, such as those encountered during extended work shifts or high-stress environments.
\end{enumerate}

\begin{figure}[th]
    \centering
    \begin{subfigure}[b]{0.49\linewidth}
        \centering
        \includegraphics[width=\linewidth]{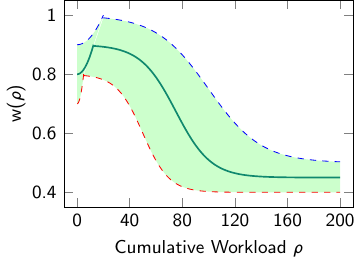}
        \vspace{-2ex}
        \caption{Cifar100}
        \label{fig:cifair100_expert}
    \end{subfigure}
    \hfill
    \begin{subfigure}[b]{0.49\linewidth}
        \centering
        \includegraphics[width=\linewidth]{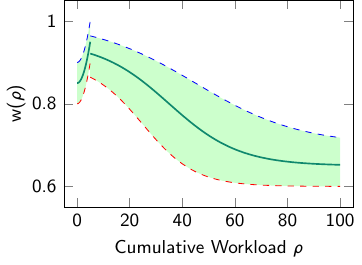}
        \vspace{-2ex}
        \caption{Chaoyang}
        \label{fig:chaoyang_expert}
    \end{subfigure}
    \\
    \vspace{1ex}
    \begin{subfigure}[b]{0.49\linewidth}
        \centering
        \includegraphics[width=\linewidth]{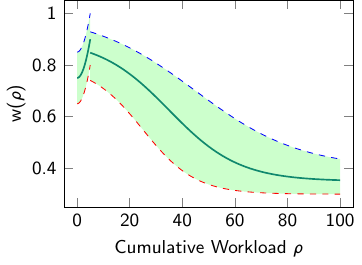}
        \vspace{-2ex}
        \caption{FLickr10K}
        \label{fig:flickr_expert}
    \end{subfigure}
    \hfill
    \begin{subfigure}[b]{0.49\linewidth}
        \centering
        \includegraphics[width=\linewidth]{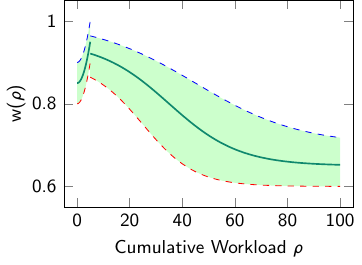}
        \vspace{-2ex}
        \caption{Micebone}
        \label{fig:micebone_expert}
    \end{subfigure}
    \caption{Human performance-Cumulative Workload curves on various datasets. The blue and red lines denote the upper and lower bound of human performance under cumulative workload accumulation. }
    \label{fig:exp_expert}
\end{figure}

For each case, we employ two distinct evaluation protocols to assess  robustness and generalisation capability. \textbf{Fine-tuning Setting:} This setting evaluates how well different approaches can adapt when they have full knowledge of the specific fatigue pattern during training. 
All training procedures remain consistent with the main experiments, but the human performance simulation for training and testing uses only the parameters from the specific case being evaluated, rather than the broader parameter ranges shown in \cref{fig:exp_expert}. \textbf{Zero-shot Setting:} This setting measures the ability to generalise to previously unseen human performance patterns. Methods use models trained on the main experiments with parameter ranges in \cref{tab:human_params_cifar100,tab:human_params_chaoyang,tab:human_params_flickr,tab:human_params_micebone} and \cref{fig:exp_expert} without additional training or adaptation for the specific case being evaluated during testing.

\section{Experiments}
\label{sec:experiments}

\subsection{Datasets}
\label{sec:data}
\textbf{Cifar100}~\citep{krizhevsky2009learning}  has 50k training images and 10k testing images, with each image belonging to one of 100 classes categorised into 20 super-classes. In addition, because about 10\% of testing images in Cifar100~\citep{krizhevsky2009learning} are duplicated or almost identical to the ones in the training set, in our training and testing, we use ciFAIR-100~\citep{barz2020we}, which replaces those duplicated images by different images belonging to the same class.

\textbf{Chaoyang}~\citep{zhu2021hard} comprises 6,160 colon slide patches categorised into four classes: \emph{normal, serrated, adenocarcinoma, and adenoma}, where each patch has \textit{three noisy labels annotated by three pathologists}. In the original Chaoyang dataset setup, the training set has patches with multi-rater noisy labels, while the testing set only contains patches that all experts agree on a single label. We assume that the majority vote forms the ground truth annotation.

\textbf{MiceBone}~\citep{schmarje2022data} has 7,240 second-harmonic generation microscopy images, with each image being annotated by one to five professional annotators, where the annotation consists of one of three possible classes: \emph{similar collagen fiber orientation, dissimilar collagen fiber orientation, and not of interest due to noise or background}. Only 8 out of 79 annotators label the whole dataset. We, therefore, use the majority vote of 8 annotators as the ground truth.

\textbf{Flickr10K}~\citep{yang2017learning} is a subset of Flickr dataset \citep{borth2013large}, in which the numbers of each class are roughly equal. It contains 10,700 images labelled with 8 commonly used emotions, including \emph{amusement, contentment, excitement, awe, anger, disgust, fear, and sadness}.

\textbf{Dataset Reshuffling.}
To ensure fair comparison across all methods, we standardise the testing episodes by reshuffling several datasets. For \textbf{Cifar100}, we retain the original test set, resulting in 50 testing episodes with 200 time steps each. For \textbf{Chaoyang}, we split the complete dataset into 4,160 training images and 2,000 testing images, yielding 20 testing episodes with 100 time steps. For \textbf{MiceBone}, we allocate 5,240 images for training and reserve the remaining 2,000 images for testing, comprising 20 episodes with 100 time steps. For \textbf{FLickr10K}, we divide the dataset into 8,700 training images and 2,000 testing images, which generates 20 testing episodes with 100 time steps.

\subsection{Architecture}
\begin{figure}[t]
    \centering
    \scalebox{0.8}{
    \begin{tikzpicture}
        \pgfplotstableread[col sep=&, row sep=\\, header=true]{
            method & time\\
            OneStage L2D & 78\\
            TwoStage L2D & 84\\
            L2D-Pop & 132 \\
            EAL2D & 126\\
            FALCON & 116\\
        } \mytable
        \pgfplotstablegetrowsof{\mytable}
        \pgfmathsetmacro{\NumRows}{\pgfplotsretval-1}  

        \hspace{-0.5em}
        \begin{axis}[
            width = 0.75\linewidth,
            xbar=0pt,
            y=1.5em,
            ytick=data,
            ymax={\NumRows + 1},
            ymin={-0.75},
            yticklabels from table={\mytable}{method},
            yticklabel style = {font=\small, align=left},
            ytick pos=left,
            xticklabel style = {font=\small},
            xlabel={Time (in minutes)},
            axis x line*=bottom,
            axis y line*=left,
            axis line style={-Latex},
            scale only axis,
            enlarge x limits=auto,
            enlarge y limits=auto,
            nodes near coords,
            every node near coord/.append style={font=\footnotesize},
            every axis plot/.append style={fill=PineGreen, draw=none}
        ]
            \addplot[] table [y expr=\NumRows - \coordindex, x=time]{\mytable};
        \end{axis}
    \end{tikzpicture}}
    \caption{Training time of FALCON and competing methods on Cifar100 (1e7 iterations).}
    \label{fig:training_time}
\end{figure}
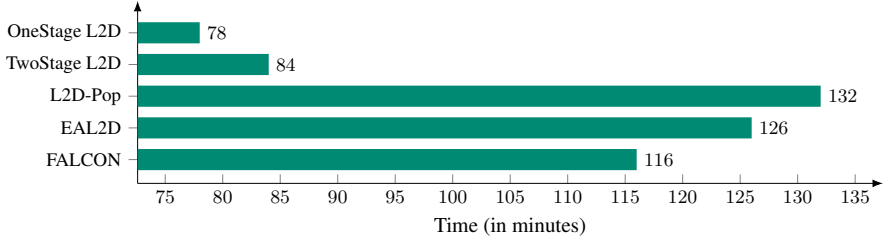

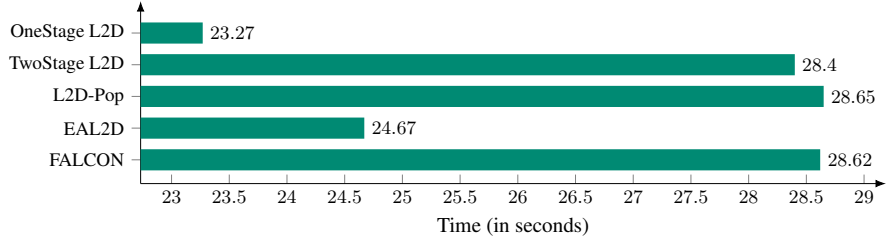
\begin{figure}[t]
    \centering
    \scalebox{0.8}{
    \begin{tikzpicture}
        \pgfplotstableread[col sep=&, row sep=\\, header=true]{
            method & time\\
            OneStage L2D & 23.27\\
            TwoStage L2D & 28.40\\
            L2D-Pop & 28.65 \\
            EAL2D & 24.67\\
            FALCON & 28.62\\
        } \mytable
        \pgfplotstablegetrowsof{\mytable}
        \pgfmathsetmacro{\NumRows}{\pgfplotsretval-1}  

        \hspace{-0.5em}
        \begin{axis}[
            width = 0.75\linewidth,
            xbar=0pt,
            y=1.5em,
            ytick=data,
            ymax={\NumRows + 1},
            ymin={-0.75},
            yticklabels from table={\mytable}{method},
            yticklabel style = {font=\small, align=left},
            ytick pos=left,
            xticklabel style = {font=\small},
            xlabel={Time (in seconds)},
            axis x line*=bottom,
            axis y line*=left,
            axis line style={-Latex},
            scale only axis,
            enlarge x limits=auto,
            enlarge y limits=auto,
            nodes near coords,
            every node near coord/.append style={font=\footnotesize},
            every axis plot/.append style={fill=PineGreen, draw=none}
        ]
            \addplot[] table [y expr=\NumRows - \coordindex, x=time]{\mytable};
        \end{axis}
    \end{tikzpicture}}
    \caption{Inference time of FALCON and competing methods on Cifar100 (50 episodes).}
    \label{fig:inference_time}
\end{figure}

All methods are implemented in Jax, a Python library that accelerates array computation and program transformation to achieve high-performance numerical computing for large-scale machine learning, while running on a single Nvidia RTX A6000. A mixed precision using \texttt{bfloat16} is applied over all methods and datasets to speed up the training. All AI models are trained for 300 epochs using stochastic gradient descent with a momentum of 0.9 and a learning rate of 0.01. The learning rate is decayed through a cosine decaying scheduler, and the gradient norm is clipped at the maximal of 10 for numerical stability.
For experiments performed on Cifar100 dataset, we employ PreAct-ResNet-18 and the batch size used is 256.
For other datasets, we train the AI model with a ResNet-18 using a regular CE loss minimisation with a ground truth label, while the batch size used is 256. On Cifar100 the AI model achieves 64.99\% accuracy on the testing set. The AI models on Chaoyang, Flickr10K, and Micebone datasets achieve 72.65\%, 81.35\%, 60.94\%, and 81.76\%, respectively. 
All methods are trained for 1e7 iterations. For our PPO-Lagrangian training, we use Adam optimiser and the parameters is shown in \cref{tab:ppo}. For other methods, we employ stochastic gradient descent with a momentum of 0.9, while the initial learning rate is set at 0.01 and decayed through a cosine annealing.  
Furthermore, the actor, reward and cost, critic heads in \cref{fig:architecture} consist of two-layer multi-layer perceptron (MLP), where each hidden layer has 512 nodes activated by Rectified Linear Units (ReLU). For our training, Cifar100 has 200 steps in each episode, while the other datasets have 100 steps\footnote{Cifar100 uses a larger number of episodes because it has more images than other datasets in FA-L2D.} 
The training parameters for PPO and the Lagrange multipliers are in \cref{tab:ppo}, while the parameters of mental fatigue curves \(\mathsf{w}(\rho)\) are provided in~\cref{tab:human_params_cifar100,tab:human_params_chaoyang,tab:human_params_flickr,tab:human_params_micebone} for different datasets. \cref{fig:training_time,fig:inference_time} show the training time for 1e7 iterations and testing time for 50 episodes on Cifar100 dataset across all methods.

\begin{table}[th]
    \centering
    \caption{PPO parameters}
    \begin{tabular}{lc}
    \toprule
    \textbf{Params}            & \textbf{Value}  \\ \midrule
    \texttt{Activation}      & Relu   \\
    \texttt{Clipping\_Coefficient} \(\epsilon\)      & 0.2    \\
    \texttt{Entropy\_Coefficient}       & 0.001  \\
    \texttt{Lagrangian\_LR}      & 0.035  \\
    \texttt{Lagrangian\_INIT \(\lambda\)}    & 0.001  \\
    \texttt{GAE\_LAMBDA}     & 0.95   \\
    \texttt{Discount Factor \(\gamma\)}           & 0.99   \\
    \texttt{LR}              & 0.0004 \\
    \texttt{LR\_WARMUP}      & 0.01   \\
    \texttt{UPDATE\_EPOCHS}  & 4      \\
    \texttt{Value Function Weight}        & 0.5    \\
    \texttt{Maximum Gradient Norm}        & 0.5    \\
    \texttt{S5 Layers}         & 4    \\
    \texttt{S5 Hidden Size}         & 512    \\
    \texttt{FC\_DIM}     & 512    \\  \bottomrule
    \end{tabular}
    
    \label{tab:ppo}
\end{table}

\begin{table}[th]
\centering
\caption{The range of parameters of human performance variation in \cref{eq:W_H} on Cifar100 dataset.}
\begin{tabular}{lcc}
\toprule
Params & Range & Description \\ \midrule
\(w_{0}\)          & \(\mathcal{U}(0.7, 0.9)\)      & initial performance            \\
\(w_\text{base}\)          & \(\mathcal{U}(0.4, 0.5)\)      & minimum performance            \\
\(w_\text{peak}\)          & \(\mathcal{U}(0.8, 1.0)\)      & maximum performance            \\
\(\hat \rho\)          & \(\mathcal{U}(0.025, 0.1)\)       & relative workload at the peak performance             \\
\(\bar{\rho}\)          & \(\mathcal{U}(0.25, 0.5)\)      & relative workload at the inflection point of the decay phase            \\
\(k\)          & \(\mathcal{U}(0.05, 0.1)\)      & steepness of performance decline            \\ \bottomrule
\end{tabular}
\label{tab:human_params_cifar100}
\end{table}

\begin{table}[th]
\centering
\caption{The range of parameters of human performance variation in \cref{eq:W_H} on Chaoyang dataset.}
\begin{tabular}{lcc}
\toprule
Params & Range & Description \\ \midrule
\(w_{0}\)          & \(\mathcal{U}(0.8, 0.9)\)      & initial performance            \\
\(w_\text{base}\)          & \(\mathcal{U}(0.6, 0.7)\)      & minimum performance            \\
\(w_\text{peak}\)          & \(\mathcal{U}(0.9, 1.0)\)      & maximum performance            \\
\(\hat \rho\)          & \(\mathcal{U}(0.025, 0.1)\)       & relative workload at the peak performance            \\
\(\bar{\rho}\)          & \(\mathcal{U}(0.25, 0.5)\)      & relative workload at the inflection point of the decay phase            \\
\(k\)          & \(\mathcal{U}(0.05, 0.1)\)      & steepness of performance decline            \\ \bottomrule
\end{tabular}
\label{tab:human_params_chaoyang}
\end{table}

\begin{table}[th]
\centering
\caption{The range of parameters of human performance variation in \cref{eq:W_H} on FLickr10K dataset.}
\begin{tabular}{lcc}
\toprule
Params & Range & Description \\ \midrule
\(w_{0}\)          & \(\mathcal{U}(0.65, 0.9)\)      & initial performance            \\
\(w_\text{base}\)          & \(\mathcal{U}(0.3, 0.4)\)      & minimum performance            \\
\(w_\text{peak}\)          & \(\mathcal{U}(0.8, 1.0)\)      & maximum performance            \\
\(\hat \rho\)          & \(\mathcal{U}(0.025, 0.1)\)       & relative workload at the peak performance             \\
\(\bar{\rho}\)          & \(\mathcal{U}(0.25, 0.5)\)      & relative workload at the inflection point of the decay phase            \\
\(k\)          & \(\mathcal{U}(0.05, 0.1)\)      & steepness of performance decline            \\ \bottomrule
\end{tabular}
\label{tab:human_params_flickr}
\end{table}

\begin{table}[th]
\centering
\caption{The range of parameters of human performance variation in \cref{eq:W_H} on Micebone dataset.}
\begin{tabular}{lcc}
\toprule
Params & Range & Description \\ \midrule
\(w_{0}\)          & \(\mathcal{U}(0.8, 0.9)\)      & initial performance            \\
\(w_\text{base}\)          & \(\mathcal{U}(0.6, 0.7)\)      & minimum performance            \\
\(w_\text{peak}\)          & \(\mathcal{U}(0.9, 1.0)\)      & maximum performance            \\
\(\hat \rho\)          & \(\mathcal{U}(0.025, 0.1)\)       & relative workload at the peak performance             \\
\(\bar{\rho}\)          & \(\mathcal{U}(0.25, 0.5)\)      & relative workload at the inflection point of the decay phase            \\
\(k\)          & \(\mathcal{U}(0.05, 0.1)\)      & steepness of performance decline            \\ \bottomrule
\end{tabular}
\label{tab:human_params_micebone}
\end{table}

\subsection{Baselines and SOTAs} 
We compare FALCON with SOTA L2D methods, such as one-stage L2D~\citep{consistentest_Mozannar2020}, two-stage L2D~\citep{madras2018predict}, L2D-Pop~\citep{tailor2024learning}, and EA-L2D~\citep{strong2025expert}. During the training phase, conventional L2D methods require all human predictions for backpropagation. Therefore, to train these models fairly, we simulate human experts by randomly generating a complete performance variation sequence for each training trajectory, where each training batch contains full trajectories and provides models access to all human predictions. For L2D-Pop and EA-L2D, we follow its prescribed methodology by training it on 16 randomly generated human expert simulations per epoch. In the testing phase, after the model makes its deferral decisions, the accuracy is calculated by incorporating the simulated human's predictions for all deferred instances. Note that we control the budget of all static L2D methods by the penalty constraint optimisation from~\citep{zhang2026coverage}. 

\begin{figure}[ht]
    \centering
    \vspace{-2ex}
    \includegraphics[width=\linewidth]{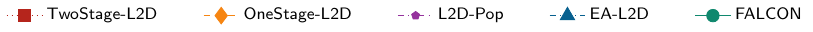}\\
    \begin{subfigure}[b]{0.49\linewidth}
        \centering
        \includegraphics[width=\linewidth]{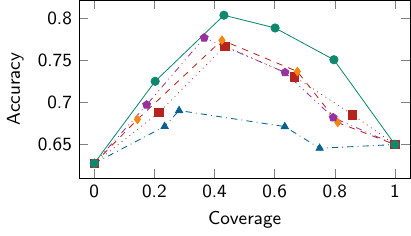}
        \vspace{-2ex}
        \caption{Cifar100}
        \label{fig:cifair100}
    \end{subfigure}
    \hfill
    \begin{subfigure}[b]{0.49\linewidth}
        \centering
        \includegraphics[width=\linewidth]{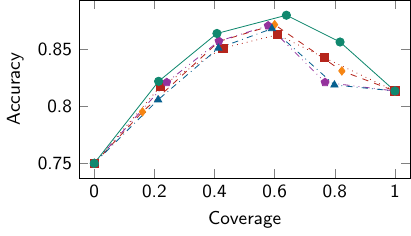}
        \vspace{-2ex}
        \caption{Chaoyang}
        \label{fig:chaoyang}
    \end{subfigure}
    \\
    \vspace{1ex}
    \begin{subfigure}[b]{0.49\linewidth}
        \centering
        \includegraphics[width=\linewidth]{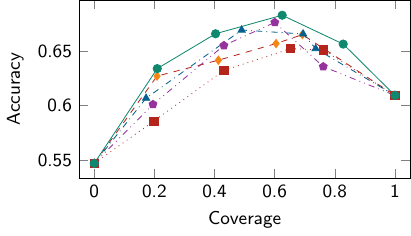}
        \vspace{-2ex}
        \caption{FLickr10K}
        \label{fig:flickr}
    \end{subfigure}
    \hfill
    \begin{subfigure}[b]{0.49\linewidth}
        \centering
        \includegraphics[width=\linewidth]{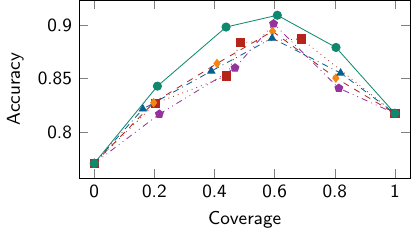}
        \vspace{-2ex}
        \caption{Micebone}
        \label{fig:micebone}
    \end{subfigure}
    \vspace{-1ex}
    \caption{Accuracy-Coverage curves of several L2D strategies and FALCON on various datasets.}
    \label{fig:exp}
    \vspace{-1ex}
\end{figure}

\subsection{Comparison with Baselines and SOTAs} 
We report the \textit{accuracy-coverage curves} of several L2D strategies and our proposed FALCON across the FA-L2D benchmark datasets in~\cref{fig:exp}. 
In general, FALCON outperforms all competing methods at every coverage level in all benchmarks. TwoStage-L2D achieves better performance at high coverage but worse than others at low coverages.
On datasets with a small number of classes, (e.g. Chaoyang, Micebone), L2D-Pop and EA-L2D show small improvements over simpler OneStage and TwoStage L2D models. This suggests that their learned human representation is weak in these scenarios. In contrast, FALCON maintains a remarkable performance advantage, especially when coverage is large, highlighting its capabilities. 
On the FLickr10K dataset, L2D-Pop and EA-L2D outperform the simpler baselines at mid-range budget levels. This indicates that they can capture an average representation of expert performance. However, FALCON's strength lies in its ability to adapt to a dynamic environment and unseen expert behaviours, rather than relying on a simple average. 
EA-L2D performs worse than other methods on  datasets with a large number of classes (e.g., Cifar100), because the counting-based prior for expert accuracy cannot scale effectively. When the number of classes increases, the gating function will be biased to the classifier. Although L2D-Pop achieves higher performance than other baselines, FALCON achieves the best results.
Regarding the AUACC results in \cref{tab:aucc} of Appendix, FALCON shows better results than all other methods for all datasets. It is worth noting the superior performances, particularly on Cifar100 and MiceBone.  All other methods perform competitively against each other, except for EA-L2D that shows poor performance on Cifar100.

\begin{table}[]
    \centering
    \caption{Quantitative comparison in terms of AUACC \((\times100)\)~\citep{nadeem2009accuracy} of the SOTA L2D~\citep{consistentest_Mozannar2020,madras2018predict,tailor2024learning,strong2025expert} on the L2D datasets. The results consist of the mean and standard deviations obtained from three experiments using models trained with different random seeds. The best result per benchmark is marked in bold.}
    \begin{tabular}{lcccc}
        \toprule
                     & Cifar100 & Chaoyang & FLickr10K & MiceBone \\ \midrule
        OneStage L2D & 70.87$\pm$0.13    & 83.24$\pm$0.14  & 63.06$\pm$0.12 & 84.61$\pm$0.12   \\
        TwoStage L2D & 70.50$\pm$0.15    & 83.15$\pm$0.08  & 61.77$\pm$0.13 & 84.58$\pm$0.15   \\
        L2D-Pop      & 71.01$\pm$0.11    & 82.65$\pm$0.17  & 62.72$\pm$0.18 & 83.96$\pm$0.14       \\
        EA-L2D       & 66.26$\pm$0.39    & 82.39$\pm$0.08           & 63.26$\pm$0.23 & 84.59$\pm$0.12          \\
        \rowcolor{gray!25} Ours         & \textbf{74.01$\pm$0.09}    & \textbf{84.13$\pm$0.11}   & \textbf{64.40$\pm$0.08} & \textbf{86.08$\pm$0.13}   \\ \bottomrule
    \end{tabular}
    \label{tab:aucc}
\end{table}

\begin{figure}[ht]
    \centering
    \includegraphics[width=\linewidth]{results/legend}\\
    \begin{subfigure}[b]{0.3\linewidth}
        \centering
        \includegraphics[width=\linewidth]{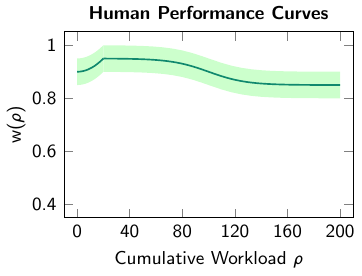}
        \caption{}
        \label{fig:case_1_curve}
    \end{subfigure}
    \hfill
    \begin{subfigure}[b]{0.3\linewidth}
        \centering
        \includegraphics[width=\linewidth]{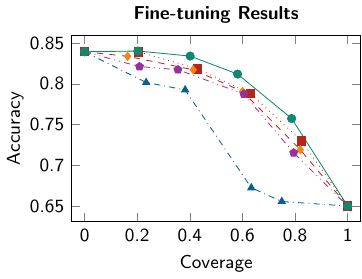}
        \caption{}
        \label{fig:case_1_results}
    \end{subfigure}
    \hfill
    \begin{subfigure}[b]{0.3\linewidth}
        \centering
        \includegraphics[width=\linewidth]{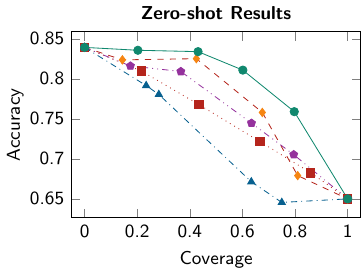}
        \caption{}
        \label{fig:0_shot_case_1_results}
    \end{subfigure}
    \\
    \begin{subfigure}[b]{0.3\linewidth}
        \centering
        \includegraphics[width=\linewidth]{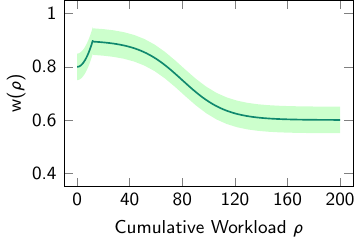}
        \caption{}
        \label{fig:case_2_curve}
    \end{subfigure}
    \hfill
    \begin{subfigure}[b]{0.3\linewidth}
        \centering
        \includegraphics[width=\linewidth]{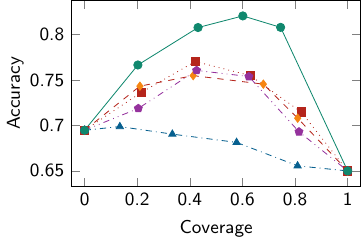}
        \caption{}
        \label{fig:case_2_results}
    \end{subfigure}
    \hfill
    \begin{subfigure}[b]{0.3\linewidth}
        \centering
        \includegraphics[width=\linewidth]{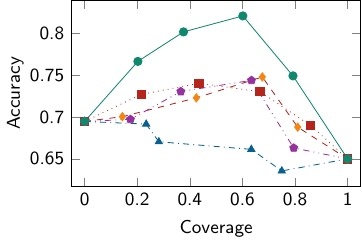}
        \caption{}
        \label{fig:0_shot_case_2_results}
    \end{subfigure}
    \\
    \begin{subfigure}[b]{0.3\linewidth}
        \centering
        \includegraphics[width=\linewidth]{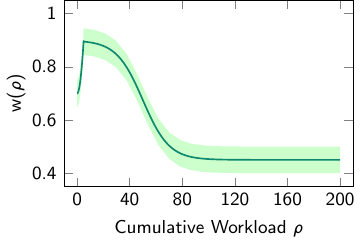}
        \caption{}
        \label{fig:case_3_curve}
    \end{subfigure}
    \hfill
    \begin{subfigure}[b]{0.3\linewidth}
        \centering
        \includegraphics[width=\linewidth]{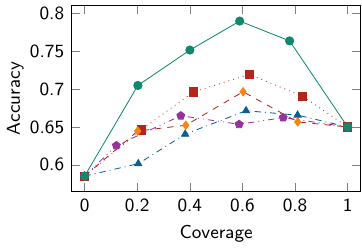}
        \caption{}
        \label{fig:case_3_results}
    \end{subfigure}
    \hfill
    \begin{subfigure}[b]{0.3\linewidth}
        \centering
        \includegraphics[width=\linewidth]{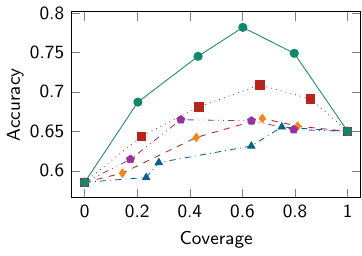}
        \caption{}
        \label{fig:0_shot_case_3_results}
    \end{subfigure}
    \\
    \label{fig:diff_human}
    \caption{Different human performance during testing (left column) and corresponding results with fine-tuning (middle column) and zero-shot (right column) testing on Cifar100 across all methods.}
\end{figure}

\begin{table}[]
\centering
\caption{Quantitative comparison in terms of the Area Under Accuracy-Coverage Curve (AUACC) \((\times100)\)~\citep{nadeem2009accuracy} of the SOTA L2D~\citep{consistentest_Mozannar2020,madras2018predict,tailor2024learning,strong2025expert} with three different human performance curves on the Cifar100 dataset. The results consist of the mean value obtained from three
experiments using models trained with different random seeds. The best result per benchmark is marked in bold.}
\resizebox{\linewidth}{!}{
\begin{tabular}{lcccccc}
\toprule
             & \multicolumn{2}{c}{\text{Sustained High Performance}}                                    & \multicolumn{2}{c}{\text{Normal Fatigue}}                & \multicolumn{2}{c}{\text{Rapid Fatigue}}                \\
             & \multicolumn{1}{l}{Fine-tuning} & Zero-shot & \multicolumn{1}{l}{Fine-tuning} & Zero-shot & \multicolumn{1}{l}{Fine-tuning} & Zero-shot \\ \midrule
OneStage L2D & 78.23$\pm$0.17   & 77.25$\pm$0.13  & 72.67$\pm$0.10   & 70.83$\pm$0.16   & 66.75$\pm$0.09    & 63.85$\pm$0.11 \\
TwoStage L2D & 78.78$\pm$0.14   & 75.22$\pm$0.16  & 73.10$\pm$0.09   & 71.56$\pm$0.07   & 67.36$\pm$0.14    & 67.49$\pm$0.13 \\
L2D-Pop      & 77.63$\pm$0.08   & 76.38$\pm$0.12  & 72.07$\pm$0.11   & 70.35$\pm$0.13   & 64.82$\pm$0.12    & 64.40$\pm$0.08 \\
EA-L2D       & 73.46$\pm$0.10   & 72.23$\pm$0.14  & 67.87$\pm$0.15   & 66.51$\pm$0.12   & 63.97$\pm$0.17    & 62.31$\pm$0.14 \\
\rowcolor{gray!25} Ours         & \textbf{79.58$\pm$0.10}   & \textbf{79.70$\pm$0.12}  &  \textbf{76.93$\pm$0.07}  & \textbf{76.20$\pm$0.09}  & \textbf{72.36$\pm$0.15}  & \textbf{71.68$\pm$0.07} \\ \bottomrule
\end{tabular}}
\label{tab:case_aucc}
\end{table}

\subsection{Robustness of L2D Methods Under FA-L2D Parameter Variations}
To evaluate the robustness of L2D methods to varying parameters of the FA-L2D benchmark, we test the methods with the ablation settings in \cref{sec:FA_L2D_Benchmark}, as illustrated in \cref{fig:case_1_curve,fig:case_2_curve,fig:case_3_curve} (above) and \cref{tab:case_aucc} (in Appendix). We evaluated each method's ability to adjust its deferral strategy under fine-tuning settings in \cref{fig:case_1_results,fig:case_2_results,fig:case_3_results} and zero-shot settings in \cref{fig:0_shot_case_1_results,fig:0_shot_case_2_results,fig:0_shot_case_3_results}.

\emph{Sustained High Performance.} 
In this scenario, the human expert's accuracy remains high, staying above 80\% for the duration of the task in \cref{fig:case_1_curve}. The results in \cref{fig:case_1_results,fig:0_shot_case_1_results} show that FALCON consistently achieves the highest accuracy. Other methods achieve similar performance in \cref{fig:case_1_curve}, which indicates their advantages in standard L2D setting. In \cref{fig:0_shot_case_1_results}, FALCON significantly exceeds that of EA-L2D and L2D-Pop, which struggle to effectively cooperate with a strong human expert. OneStage-L2D performs better than EA-L2D and L2D-Pop, suggesting that these methods cannot learn efficient dynamic human presentation.

\begin{figure}[]
    \centering
    \includegraphics[width=0.48\linewidth]{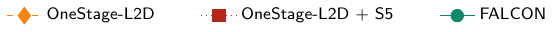}\\
    \includegraphics[width=0.48\linewidth]{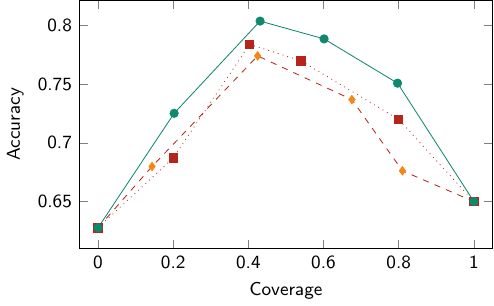}
    \caption{Accuracy-Coverage curves of CMDP ablation on Cifar100 dataset.}
    \label{fig:s5_ablation}
\end{figure}
    
\begin{figure}[]
    \centering
    
    \begin{subfigure}[b]{0.49\linewidth}
        \centering
        \includegraphics[width=\linewidth]{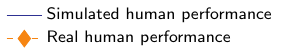}\\\includegraphics[width=\linewidth]{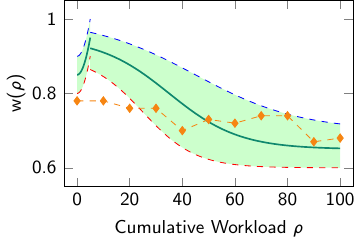}
        \label{fig:mimic_human_chaoyang}
        \caption{Simulated vs. real human performance}
    \end{subfigure}
    \hfill
    \begin{subfigure}[b]{0.49\linewidth}
        \centering
        \includegraphics[width=\linewidth]{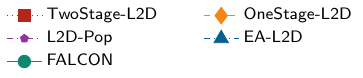}\\
        \includegraphics[width=\linewidth]{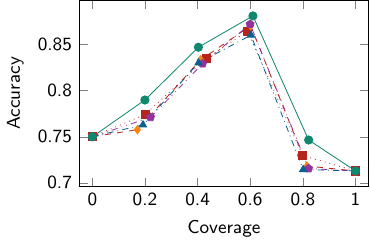}
        \label{fig:human_chaoyang_result}
        \caption{Zero-shot with real human data}
    \end{subfigure}
    \caption{Validation against clinical human performance data. (a) Comparison between simulated human performance in FALCON and real-world human data~\citep{cowley1997time}. The orange line represents the radiologist's correct recall rate, which shows a significant decline from 78\% to 66\% after the vigilance threshold. The green shaded area indicates the range of our simulated experts on the Chaoyang dataset, demonstrating high ecological validity. (b) Zero-shot testing results on the Chaoyang dataset.}
    \label{fig:mimic_human_chaoyang_performance}
\end{figure}

\emph{Normal Fatigue.} In this case, human performance peaks around 40 steps before gradually decreasing in \cref{fig:case_2_curve}. 
By letting the model 
learn the relatively slow human performance variation, performance in \cref{fig:case_2_results} is better than in \cref{fig:0_shot_case_2_results}, but also lag behind FALCON. In \cref{fig:0_shot_case_2_results}, the performance of all the methods is close to the results in \cref{fig:cifair100}.

\emph{Rapid Fatigue.} Human performance decreases from above 90\% to below 50\% in the first 80 time steps (\cref{fig:case_3_curve}). This challenging condition highlights the robustness of our approach. The results in \cref{fig:case_3_results,fig:0_shot_case_3_results} show that FALCON maintains high precision by correctly identifying human unreliability and adjusting its deferral strategy. In contrast, all other methods suffer a significant performance drop. While TwoStage-L2D is the best of the baselines, its accuracy still lags behind FALCON at all coverage levels.

\begin{remark}
    In the almost static L2D setting (e.g., sustained high-performance humans), it becomes harder to appreciate the value of L2D methods (see \cref{fig:case_1_curve,fig:case_1_results,fig:0_shot_case_1_results}): L2D does not exhibit improved performance for coverages strictly between 0 and 1. By contrast, in the normal and rapid fatigue settings (\cref{fig:case_2_curve,fig:case_3_curve}), L2D methods surpass both AI-only and human-only baselines across intermediate coverage levels (\cref{fig:case_2_results,fig:0_shot_case_2_results,fig:case_3_results,fig:0_shot_case_3_results}), highlighting the effectiveness of adaptive human–AI cooperation in scenarios that more closely mirror real-world conditions.
\end{remark}

\subsection{Ablation Study of CMDP}

{To isolate the contribution of our CMDP formulation from the S5 temporal architecture, we evaluate a variant \textbf{OneStage-L2D + S5}, which augments static L2D with S5 layers but trains with the standard cross-entropy loss rather than PPO-Lagrangian. Due to the actual workload \(\rho\) is non-differentiable, we update the workload via the probability of querying humans for a fair comparison.}

{Results in~\cref{fig:s5_ablation} reveal that adding temporal memory without the CMDP formulation improves the performance at high coverage levels (i.e., when deferring less to humans), while degrades performance at low coverage levels (i.e., when deferring more to humans). In contrast, FALCON consistently outperforms both baselines across all coverage levels. This demonstrates that the performance gains arise from the principled integration of fatigue-aware state dynamics with temporal memory.}

{\subsection{Real-world Human Studies}}
To validate our model against real human behaviour, we use the recall vs. workload curve reported in mammographic film reading~\citep{cowley1997time}, where radiologist recall declines from 78\% to 66\% over 100 continuous readings without warm-up phase. As shown in \cref{fig:mimic_human_chaoyang_performance}(a), we map this fatigue pattern to the Chaoyang dataset, which closely matches our simulated ``Normal Fatigue'' setting. The corresponding zero‑shot results in \cref{fig:mimic_human_chaoyang_performance}(b) demonstrate that \textsc{FALCON} remains highly effective, outperforming all competing methods under this real‑world fatigue profile.
\section{Related Work}
\textbf{Learning to defer} 

learns a classifier and a rejector to decide when predictions should be deferred to a human expert~\citep{madras2018predict}. Early L2D methods largely focused on surrogate losses consistent with the Bayes-optimal classifier~\citep{consistentest_Mozannar2020,dce,cao2024defense}, but overlook settings involving diverse or multiple experts. Recent work therefore extends L2D to \emph{multiple-expert}~\citep{mao2023two,multil2d,mao2024regression,lecodu,nguyen2025probabilistic} and \emph{unseen-expert} scenarios~\citep{tailor2024learning,strong2025expert}. For example, L2D-Pop~\citep{tailor2024learning} constructs latent context representations from few-shot expert annotations, while EA-L2D~\citep{strong2025expert} derives a Bayesian estimate of each expert’s class-level performance. Sequential Learning-to-Defer~\citep{joshi2023learning} models L2D as a model-based RL problem, but focuses on environments with evolving task rules and requires batch data from human experts, making it costly for practical L2D systems.

\textbf{Human mental fatigue} 
is a critical component of non-technical skills within human factors research~\citep{casali2019rise}. Mental fatigue is a psychobiological state resulting from prolonged cognitive engagement~\citep{driskell2013stress,van2022drop}, manifesting physiologically through changes in brain activity~\citep{muller2021neural}, behaviourally through systematic declines in cognitive performance~\citep{lindner2020perceived}, and subjectively through increased perceived effort~\citep{hockey2013psychology}. Recent work also shows that mental fatigue can impair physical performance~\citep{enoka2016translating,van2017effects,marcora2009mental,dallaway2022cognitive}. Its temporal dynamics vary by task: simple repetitive or vigilance tasks often follow exponential decay~\citep{anderson2013cognitive}, and Jaber~et~al.~\citep{jaber2013incorporating} modelled fatigue–recovery cycles with exponential functions.
However, complex adaptive tasks requiring sustained cognitive engagement exhibit sigmoid performance patterns~\citep{enoka2016translating,gyles2023psychometric}.
Leppink~et~al.~\citep{leppink2019mental} observed that mental effort scales non-linearly with workload and time, with vigilance decline following non-linear patterns~\citep{mccarley2021psychometric} and cognitive load relationships exhibiting sigmoid curves~\citep{estes2015workload}.


Unlike existing L2D methods that treat human experts as static oracles, our approach explicitly models workload‑dependent performance degradation using psychologically grounded functions, enabling deferral decisions that account for the expert’s current cognitive state.

\section{Conclusion}

In this paper, we proposed FALCON to model dynamic human performance degradation due to cognitive fatigue. By formulating L2D as a CMDP with psychologically-grounded fatigue curves and PPO-Lagrangian optimisation, FALCON addresses the unrealistic assumption of static human expert performance in existing methods.
Extensive experiments on our proposed FA-L2D benchmark demonstrated that FALCON consistently outperformed SOTA L2D approaches across all coverage levels and achieved robust zero-shot generalisation to unseen expert fatigue patterns.

While FALCON captures general patterns of cognitive decline, it may not fully represent individual variation in fatigue patterns across different populations or task contexts. Also, the evaluation relies on simulated human performance rather than real human studies. FALCON assumes uniform fatigue contribution across tasks, but cognitive load varies significantly with task complexity. 
Future work will incorporate instance-dependent fatigue modelling, extending \cref{eq:sample_human_prediction,eq:cumulative_workload} to account for instance-specific cognitive load, while incorporating multi-modal sensitive fatigue indicators and real-world deployment studies.

\section{Acknowledgement}
This work has been funded by the Engineering and Physical Sciences Research Council (EPSRC) through grant EP/Y018036/1.
\bibliography{iclr2026_conference}
\bibliographystyle{unsrt}
\end{document}